# Provable Noisy Sparse Subspace Clustering using Greedy Neighbor Selection: A Coherence-Based Perspective

Jwo-Yuh Wu, Wen-Hsuan Li, Liang-Chi Huang, Yen-Ping Lin, Chun-Hung Liu, and Rung-Hung Gau

**Abstract**—Sparse subspace clustering (SSC) using greedy-based neighbor selection, such as matching pursuit (MP) and orthogonal matching pursuit (OMP), has been known as a popular computationally-efficient alternative to the conventional $\ell_1$-minimization based methods. Under deterministic bounded noise corruption, in this paper we derive coherence-based sufficient conditions guaranteeing correct neighbor identification using MP/OMP. Our analyses exploit the maximum/minimum inner product between two noisy data points subject to a known upper bound on the noise level. The obtained sufficient condition clearly reveals the impact of noise on greedy-based neighbor recovery. Specifically, it asserts that, as long as noise is sufficiently small so that the resultant perturbed residual vectors stay close to the desired subspace, both MP and OMP succeed in returning a correct neighbor subset. A striking finding is that, when the ground truth subspaces are well-separated from each other and noise is not large, MP-based iterations, while enjoying lower algorithmic complexity, yield smaller perturbation of residuals, thereby better able to identify correct neighbors and, in turn, achieving higher global data clustering accuracy. Extensive numerical experiments are used to corroborate our theoretical study.

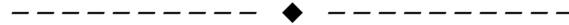

## 1 INTRODUCTION

Subspace clustering [1] is a long-standing research problem in machine learning that has found a multitude of applications in, e.g., computer vision, image processing, bioinformatics, and system theory. One popular approach to subspace clustering relies on *neighbor identification* for constructing a similarity graph of the given data set, followed by *spectral clustering* [2] for subspace data segmentation (see the tutorial introduction in [1], and a recent work [3]). Under this algorithm flow, accurate neighbor identification is rather crucial for achieving high global data clustering accuracy. Inspired by the state-of-the-arts compressive sensing (CS) and sparse representation techniques [4-7], sparse subspace clustering (SSC) [8] has received considerable attention in the recent years. The distinctive feature of SSC lies in conducting neighbor identification by means of sparse linear combination of the data points [8]. Existing such sparse regression schemes commonly resort to the $\ell_1$-minimization technique or variants thereof [8-12], which would be computationally demanding in practical applications. Low algorithmic-complexity alternatives using greedy-based neighbor selection, such as matching pursuit (MP) or orthogonal matching pursuit (OMP) [4-5, 13-16], have then been proposed [17]. Thanks to the fruitful analytical tools in CS, mathematical performance guarantees of SSC have been addressed in the literature, many among which investigated sufficient conditions guaranteeing the so-called *subspace detection property* (SDP) [9], i.e., sparse regression surely returns a neighbor subset from the correct cluster. Such related works include, e.g., [8-12] regarding the $\ell_1$-minimization solutions, and [17-19] for greedy search.

For OMP-based neighbor recovery in the noiseless case, coherence-based sufficient conditions ensuring SDP have been investigated in [17-18]. In the presence of Gaussian noise corruption, performance guarantee of MP/OMP characterized using subspace affinity was recently reported in [19]. It is worth noting that coherence and subspace affinity based conditions can offer insights into neighbor identification from quite different perspectives. The former typically involves certain *local* geometric features, such as coherence between distinct clusters and in-radius of the convex hull of individual clusters, which depend entirely on the given data set. On the contrary, the latter often calls for *global* geometric properties like subspace orientation, dimensions, and sample density (i.e., the number of data samples within each subspace/cluster), mostly related to the underlying ground truth subspaces.

————————————————
- *J. Y. Wu, W. H. Li, L. C. Huang, Y. P. Lin, and R. H. Gau are with the Institute of Communications Engineering, the Department of Electrical and Computer Engineering, and College of Electrical Engineering, National Chiao-Tung University, 1001, Ta-Hsueh Road, Hsinchu, Taiwan. Emails: jywu@cc.nctu.edu.tw; fiend_danger@hotmail.com; uuranus001@gmail.com; b8974691234@gmail.com; runghunggau@g2.nctu.edu.tw;*
- *C. H. Liu is with the Department of Electrical and Computer Engineering, Mississippi State University, USA. E-mail: chliu@ece.msstate.edu.*

In this paper we develop coherence-based performance guarantees for MP/OMP-based noisy SSC. Contrary to the probabilistic framework in [19], our study is in a *deterministic* setting, in which the noise is subject to nothing but a known upper bound on its power. The adopted deterministic formulation can offer kind of a *worst-case* solution, which is not only of theoretical interest per se but also insightful especially when the true noise/data distributions are hard (or even impossible) to know in practice. Under the deterministic bounded noise assumption, we first analytically characterize the maximal/minimal inner product between a pair of noisy data points. Afterwards, we derive sufficient conditions ensuring the SDP when MP/OMP is utilized for neighbor recovery. The obtained analytical results reveal the impact of noise corruption on greedy search of neighbors. Due to noise perturbation, the residual vector in each iteration deviates from the true subspace. Our analytic conditions show that, as long as noise and the incurred deviation of the residual, measured by the angle of deviation (AoD), are sufficiently small, the SDP is fulfilled, i.e., both MP and OMP return a correct neighbor in each iteration. In addition, it can be deduced from our analytic conditions that, when the subspaces are well-separated from each other and noise is not large, the residual vectors of MP tend to yield smaller AoD as compared to OMP. Hence, MP can achieve a better global data clustering performance, even with reduced algorithmic complexity. Extensive experimental results using both synthetic and real human face data are provided to validate our theoretical analyses. Overall, our study extends the works [17-18] to case with deterministic bounded noise corruption. It presents an original contribution to performance guarantees of noisy SSC using greedy neighbor selection, by offering a better fundamental understanding of how the noise impacts the MP/OMP-based greedy neighbor identification.

The rest of this paper is organized as follows. Section II introduces the system model and some basic assumptions. Section III presents the main results, with detailed mathematical proofs given in Section IV. Section V presents the experimental results. Finally, Section VI concludes this paper.

## 2 SIGNAL MODEL

Consider a family of $L$ subspaces $\{\mathcal{S}_1, \cdots, \mathcal{S}_L\}$ of $\mathbb{R}^n$ such that $\mathcal{S}_l$ is of dimension $0 < d_l < n$, $1 \leq l \leq L$. Let $\mathcal{X} = \{\mathbf{x}_1, \cdots, \mathbf{x}_N\}$ be a set of $N$ unit-norm vectors that obeys the ground truth being a disjoint union as

$$\mathcal{X} = \mathcal{X}_1 \cup \mathcal{X}_2 \cup \cdots \cup \mathcal{X}_L, \quad (1)$$

where $\mathcal{X}_l \subset \mathcal{S}_l$ is a finite subset consisting of $N_l > 0$ vectors; thus, $N_1 + \cdots + N_L = N$. Let $\mathcal{Y} = \{\mathbf{y}_1, \cdots, \mathbf{y}_N\}$ be the observed data set under additive noise corruption such that

$$\mathbf{y}_i = \mathbf{x}_i + \mathbf{e}_i, \ 1 \leq i \leq N, \quad (2)$$

TABLE I
MP AND OMP NEIGHBOR IDENTIFICATION ALGORITHMS

| MP algorithm | OMP algorithm |
|---|---|
| **Input:** $\mathbf{Y} = [\mathbf{y}_1 \ \cdots \ \mathbf{y}_N]$, $\mathbf{y}_i$, $m_{\max}$, $\tau$ | |
| **Initialize** residual $\mathbf{r}_0^{(i)} = \mathbf{y}_i$, $m = 0$, $\mathbf{c}_i^* = \mathbf{0}$. | **Initialize** residual $\mathbf{r}_0^{(i)} = \mathbf{y}_i$, $m = 0$, index set $\Lambda_0 = \emptyset$. |
| **While** $m < m_{\max}$ and $\|\mathbf{r}_0^{(i)}\|_2 > \tau$ **Do** $m = m+1$, $i_m = \underset{j \in \{1, \cdots N\} \setminus \{i\}}{\operatorname{argmax}} \left\| \mathbf{r}_{m-1}^{(i)T} \mathbf{y}_j \right\|$, $[\mathbf{c}_i^*]_{i_m} = [\mathbf{c}_i^*]_{i_m} + \left\| \mathbf{r}_{m-1}^{(i)T} \mathbf{y}_{i_m} \right\| / \|\mathbf{y}_{i_m}\|_2^2$, $\mathbf{r}_m^{(i)} = (\mathbf{I} - \mathbf{y}_{i_m}(\mathbf{y}_{i_m})^T / \|\mathbf{y}_{i_m}\|_2^2) \mathbf{r}_{m-1}^{(i)}$ **Until** Stopping criterion holds | **While** $m < m_{\max}$ and $\|\mathbf{r}_0^{(i)}\|_2 > \tau$ **Do** $m = m+1$, $i_m = \underset{j \in \{1, \cdots N\} \setminus \{\Lambda_{m-1} \cup i\}}{\operatorname{argmax}} \left\| \mathbf{r}_{m-1}^{(i)T} \mathbf{y}_j \right\|$, $\Lambda_m = \Lambda_{m-1} \cup \{i_m\}$, $\mathbf{Y}_{\Lambda_m} = [\mathbf{Y}_{\Lambda_{m-1}} \ \mathbf{y}_{i_m}]$, $\mathbf{r}_m^{(i)} = (\mathbf{I} - \mathbf{Y}_{\Lambda_m}(\mathbf{Y}_{\Lambda_m})^\dagger) \mathbf{y}_i$ **Until** Stopping criterion holds |
| **Output:** $\mathbf{c}_i^*$ | **Output:** $\mathbf{c}_i^* = \underset{\mathbf{c}:\mathrm{supp}(\mathbf{c}) \subseteq \Lambda_m}{\operatorname{argmin}} \|\mathbf{y}_i - \mathbf{Y}\mathbf{c}\|_2$ |

TABLE II
OUTLINE OF THE NOISY SSC ALGORITHM WITH MP/OMP

**Input:** observed data set $\mathcal{Y} = \{\mathbf{y}_1, \cdots, \mathbf{y}_N\}$, $m_{\max}$, $\tau$
1. For each $\mathbf{y}_i$, obtain the coefficient vector $\mathbf{c}_i^*$ by using MP/OMP in Table I.
2. Form a similarity graph $G$ with $N$ nodes, in which the weight on the edge between the $(i,j)$ node pair equals $|c_{i,j}^*| + |c_{j,i}^*|$.
3. Apply spectral clustering [2] to the similarity graph $G$.

**Output:** Partition $\mathcal{Y} = \widehat{\mathcal{Y}}_1 \cup \cdots \cup \widehat{\mathcal{Y}}_{\hat{L}}$

where $\mathbf{e}_i \in \mathbb{R}^n$ is the noise vector. On account of (1) and (2), $\mathcal{Y}$ admits the corresponding ground truth partition

$$\mathcal{Y} = \mathcal{Y}_1 \cup \mathcal{Y}_2 \cup \cdots \cup \mathcal{Y}_L, \quad (3)$$

where $\mathcal{Y}_l$ is the cluster $\mathcal{X}_l$ subject to noise. Given $\mathcal{Y}$ and without knowing the underlying subspaces $\{\mathcal{S}_l\}_{1 \leq l \leq L}$, their dimensions $\{d_l\}_{1 \leq l \leq L}$, and the number $L$ of subspaces, the task of subspace clustering is to uncover the partition (3). In this paper we focus on SSC using MP/OMP for neighbor identification (see Tables I and II for an algorithm outline). Our main purpose is to develop coherence-based sufficient conditions ensuring the subspace detection property (SDP) [9].

**Definition 2.1 (Subspace Detection Property (SDP)).** *Consider a data point $\mathbf{y} \in \mathcal{Y}_k$ with the neighbor index set $\Lambda$ determined by either MP or OMP. We say SDP is achieved if $\mathbf{y}_i \in \mathcal{Y}_k$ for all $i \in \Lambda$.* □

The following assumption is made throughout this paper.

**Assumption 2.2.** *For each $1 \leq i \leq N$, the noise vector $\mathbf{e}_i$ is bounded with $\|\mathbf{e}_i\|_2 \leq \varepsilon$ for some known $\varepsilon > 0$.* □

Table III summarizes the notation used in this paper. Of particular interest are the worst-case inter-cluster coherence $\mu_c(\mathcal{X}_k)$, which gauges the closeness of cluster $\mathcal{X}_k$ to its nearest cluster, and the in-radius $r_k$, which measures how uniform the data points are distributed inside $\mathcal{X}_k$. A smaller $\mu_c(\mathcal{X}_k)$ means that $\mathcal{X}_k$ is well-separated from all the other clusters, whereas a large $r_k$ implies that the data points are well spread in $\mathcal{X}_k$ and in general promotes SDP (see the discussions in [9]).

TABLE III
SUMMARY OF NOTATION

| Notation | Description |
|---|---|
| $\|\cdot\|_2$ | Euclidean vector two-norm |
| $\mu_c(\mathcal{X}_k, \mathcal{X}_l) \triangleq \max_{\mathbf{u} \in \mathcal{X}_k, \mathbf{v} \in \mathcal{X}_l} |\mathbf{u}^T \mathbf{v}|$ | mutual coherence between clusters $\mathcal{X}_k$ and $\mathcal{X}_l$ |
| $\mu_c(\mathcal{X}_k)$ | $\max_{k \neq l} \mu_c(\mathcal{X}_k, \mathcal{X}_l)$ |
| $\theta_{kl} \triangleq \arccos\left( \max_{\mathbf{u} \in \mathcal{S}_k, \mathbf{v} \in \mathcal{S}_l} \frac{|\mathbf{u}^T \mathbf{v}|}{\|\mathbf{u}\|_2 \|\mathbf{v}\|_2} \right)$ | minimal angle between the two subspaces $\mathcal{S}_k$ and $\mathcal{S}_l$ |
| $\theta_k$ | $\min_{1 \leq l \leq L, k \neq l} \theta_{kl}$ |
| $\mathbf{X}_{-i}^{(k)} \in \mathbb{R}^{n \times (N_k - 1)}$ | matrix whose columns consist of all except the $i$th vectors in $\mathcal{X}_k$ |
| $\mathcal{P}(\mathbf{X}_{-i}^{(k)})$ | symmetric convex hull of columns of $\mathbf{X}_{-i}^{(k)}$ |
| $r(\mathcal{P}(\mathbf{X}_{-i}^{(k)}))$ | in-radius [9] of the convex body $\mathcal{P}(\mathbf{X}_{-i}^{(k)})$ |
| $r_k$ | $\min_{i: \mathbf{x}_i \in \mathcal{X}_k} r(\mathcal{P}(\mathbf{X}_{-i}^{(k)}))$ |

## 3 MAIN RESULTS

In the absence of noise (i.e., $\mathcal{Y}_l = \mathcal{X}_l$ for all $1 \leq l \leq L$), a well-known coherence-based performance guarantee for OMP-SSC has been reported in [17], asserting that SDP holds if ($\mathcal{S}_k$ is assumed to be the desired subspace)

$$\mu_c(\mathcal{X}_k) < r_k - \frac{2\sqrt{1-\gamma_k^2}}{\sqrt[4]{12}} \cos \theta_k. \qquad (4)$$

Later, an improved coherence-based sufficient condition was reported in [18], saying SDP is fulfilled if

$$\max_{l \neq k} \mu_c(\mathcal{W}_k, \mathcal{X}_l) < r_k, \qquad (5)$$

where $\mathcal{W}_k \subset \mathbb{R}^n$ consists of all normalized residual vectors (during the conduction of OMP search) for all data points in $\mathcal{S}_k$. Under the bounded noise assumption, below we present new coherence-based performance guarantees for both MP and OMP. To formalize matters, for a fixed $\varphi \in [0, \pi/2]$ we define the function $f_\varphi : [0, 2\pi] \to \mathbb{R}$ as

$$f_\varphi(\theta) \triangleq 2\varepsilon \cos((\varphi/2) + \theta) + \varepsilon^2 \cos(2\theta), \qquad (6)$$

where $\varepsilon > 0$ is the noise level as in Assumption 2.2. Our analyses are built on the following technical lemma, which pins down the maximal/minimal inner product of two unit-norm vectors under bounded noise perturbation (proof referred to Section IV).

**Lemma 3.1.** Let $\mathbf{x}_i, \mathbf{x}_j \in \mathbb{R}^n$ be a given (but arbitrary) pair of unit-norm noiseless data vectors such that $0 < \cos^{-1}(\mathbf{x}_i^T \mathbf{x}_j) < \pi/2$. We have

$$\underset{\|\mathbf{e}_i\|_2 \leq \varepsilon, \|\mathbf{e}_j\|_2 \leq \varepsilon}{\text{Maximum}} (\mathbf{x}_i + \mathbf{e}_i)^T (\mathbf{x}_j + \mathbf{e}_j)$$
$$= \mathbf{x}_i^T \mathbf{x}_j + \underset{\theta \in [0, 2\pi]}{\text{Maximum}} f_{\cos^{-1}(\mathbf{x}_i^T \mathbf{x}_j)}(\theta), \qquad (7)$$

and

$$\underset{\|\mathbf{e}_i\|_2 \leq \varepsilon, \|\mathbf{e}_j\|_2 \leq \varepsilon}{\text{Minimum}} (\mathbf{x}_i + \mathbf{e}_i)^T (\mathbf{x}_j + \mathbf{e}_j)$$
$$= \mathbf{x}_i^T \mathbf{x}_j + \underset{\theta \in [0, 2\pi]}{\text{Minimum}} f_{\cos^{-1}(\mathbf{x}_i^T \mathbf{x}_j)}(\theta), \qquad (8)$$

where $f_{\cos^{-1}(\mathbf{x}_i^T \mathbf{x}_j)}$ is defined in (6). □

Based on Lemma 3.1, the main result of this paper is shown in the next theorem (proof referred to Section IV).

**Theorem 3.2.** Both MP and OMP return a correct neighbor subset for the data vector $\mathbf{y}_i \in \mathcal{Y}_k$ if

$$\mu_c(\mathcal{X}_k) < r_k - \varepsilon \left\{ \max_{\theta \in [0, 2\pi]} f_{\cos^{-1}(\mu_c(\mathcal{X}_k))}(\theta) - \min_{\theta \in [0, 2\pi]} f_{\cos^{-1}(r_k)}(\theta) \right\}$$
$$\qquad (9)$$

and, for each $m \geq 1$,

$$\cos\left(\max\{\theta_k - \phi_m^{(i)}, 0\}\right) + 2\varepsilon < \max_{\mathbf{x}_j \in \mathcal{X}_k \setminus \{\mathbf{x}_i\}} \left|\cos \angle (\mathbf{r}_m^{(i)}, \mathbf{x}_j)\right|, \quad (10)$$

in which $\mathbf{r}_m^{(i)}$ is the residual vector at the $m$th iteration with AoD defined to be

$$\phi_m^{(i)} \triangleq \tan^{-1}\left( \frac{\|(\mathbf{I} - \mathbf{P}_{\mathcal{S}_k}) \mathbf{r}_m^{(i)}\|_2}{\|\mathbf{P}_{\mathcal{S}_k} \mathbf{r}_m^{(i)}\|_2} \right), \qquad (11)$$

where $\mathbf{P}_{\mathcal{S}_k}$ is the orthogonal projection matrix onto the subspace $\mathcal{S}_k$. □

Some discussions regarding the above theorem are given below.

1. Compared to the bounds (4) and (5) for the noiseless case, in which SDP is ensured by just a single inequality, our sufficient condition in the noisy environment calls for different inequalities from iteration to iteration. To see the insights offered by Theorem 3.2 into MP/OMP-based neighbor search, we first consider the sufficient condition (9), which ensures correct neighbor recovery in the first iteration. Recall that the coherence $\mu_c(\mathcal{X}_k)$ is a worst-case measure of separation between the inter-cluster noiseless signal points, whereas the in-radius $r_k$ gauges how uniformly the signal points in the desired cluster $\mathcal{X}_k$ are spread over the subspace $\mathcal{S}_k$ [9]. Basically, the second term on the right-hand-side (RHS) of (9), i.e., $\varepsilon \{ \max_{\theta \in [0,2\pi]} f_{\cos^{-1}(\mu_c(\mathcal{X}_k))}(\theta) - \min_{\theta \in [0,2\pi]} f_{\cos^{-1}(r_k)}(\theta) \}$, can be regarded as the penalty incurred by noise at the level $\varepsilon$. Note that, as $\varepsilon$ increases, the feasible noise vector set, $\{\mathbf{e}_i, \mathbf{e}_j \in \mathbb{R}^n \mid \|\mathbf{e}_i\|_2 \leq \varepsilon, \|\mathbf{e}_j\|_2 \leq \varepsilon\}$, of the optimization problem underlying (7) and (8) is enlarged; as a result, the maximum and minimum of the noisy inner product $(\mathbf{x}_i + \mathbf{e}_i)^T(\mathbf{x}_j + \mathbf{e}_j)$ will in turn increase and decrease, respectively. This immediately implies that the aforementioned noise-induced penalty increases and, therefore, the upper bound in (9) decreases with $\varepsilon$. We thus conclude that, for a given pair of $\mu_c(\mathcal{X}_k)$ and $r_k$ associated with the data set, a stronger noise level $\varepsilon$ incurs a larger penalty and imposes a more stringent requirement on the separation between different clusters (a smaller $\mu_c(\mathcal{X}_k)$) to warrant SDP.

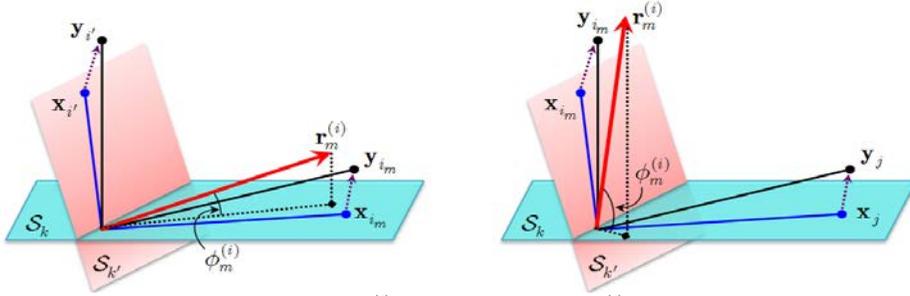

Fig. 1. Illustration of the sufficient condition (10). For small $\phi_m^{(i)}$ so that the residual $\mathbf{r}_m^{(i)}$ stays close to the desired subspace $\mathcal{S}_k$ (left figure), the data point $\mathbf{y}_{i_m}$ picked by the algorithm in the $m$th iteration is a correct neighbor. If $\phi_m^{(i)}$ is large and $\mathbf{r}_m^{(i)}$ is pushed much toward another subspace $\mathcal{S}_{k'}$, the algorithm mis-identifies a $\mathbf{y}_{i_m}$ originating from another cluster (right figure).

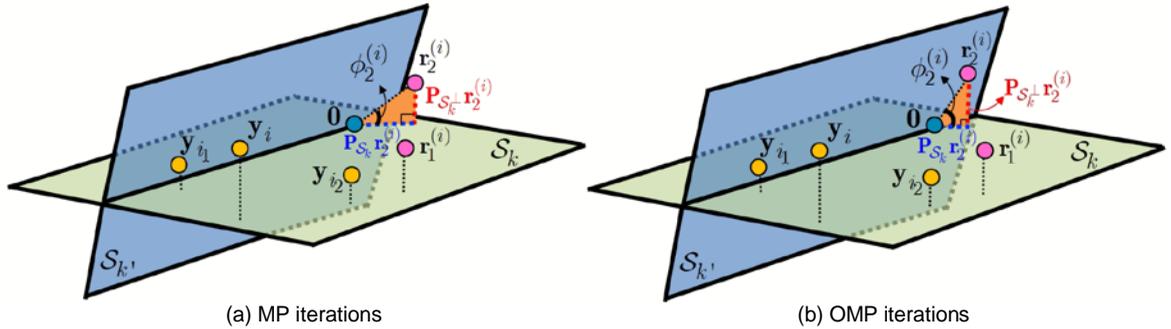

(a) MP iterations  (b) OMP iterations

Fig. 2. An illustrative example of Observation 3-1, with ambient space $\mathbb{R}^3$, two subspaces (two-dimensional) $\mathcal{S}_k$ and $\mathcal{S}_{k'}$ well-separated from each other so that SDP holds, i.e., $\{\mathbf{y}_{i_1},\mathbf{y}_{i_2}\} \subset \mathcal{S}_k$ (for ease of illustration, only the first two iterations are considered, i.e., $m=2$). The orthogonal projection $\mathbf{P}_{\mathcal{S}_k}\mathbf{r}_2^{(i)}$ in the MP case (left figure) is larger than that in the OMP case (right figure), due to joint orthogonal projection conducted by OMP. On the other hand, the orthogonal complement $\mathbf{P}_{\mathcal{S}_k^\perp}\mathbf{r}_2^{(i)}$ in both cases are very close. As a result, the AoD $\phi_2^{(i)}$ of the residual vector for MP is smaller.

2. Assume that (9) holds so that the algorithm returns a correct neighbor, say, $\mathbf{y}_{i_1} \in \mathcal{Y}_k$, in the first iteration. Let us then move on to the condition (10), which ensures correct neighbor recovery for all subsequent iterations. Due to the presence of noise, the residual vector $\mathbf{r}_m^{(i)}$ computed in the $m$th iteration ($m \geq 1$) no longer stays in the true subspace $\mathcal{S}_k$. The AoD $\phi_m^{(i)}$ in (11) then pins down how much the residual $\mathbf{r}_m^{(i)}$ deviates from the desired subspace $\mathcal{S}_k$. For the algorithm to succeed in finding a correct neighbor in the $(m+1)$th iteration, inequality (10) imposes a requirement that $\phi_m^{(i)}$ should be small enough, i.e., $\mathbf{r}_m^{(i)}$ cannot deviate much from $\mathcal{S}_k$. Indeed, note that $\theta_k$ can be regarded as the "worst-case margin" that can tolerate the deviation of $\mathbf{r}_m^{(i)}$ from $\mathcal{S}_k$. In this regard, we can think of the left-had-side (LHS) of (10) as a measure of the closeness of the residual $\mathbf{r}_m^{(i)}$ to the subspace nearest to $\mathcal{S}_k$, under the noise perturbation level $\varepsilon$. On the other hand, the RHS of (10) reflects how close $\mathbf{r}_m^{(i)}$ is to the desired cluster $\mathcal{X}_k$. Hence, if (10) is true, $\mathbf{r}_m^{(i)}$ is guaranteed to stay closer to the ground truth cluster $\mathcal{X}_k$ and, therefore, the algorithm can return a correct neighbor from the desired cluster $\mathcal{Y}_k$ in the $(m+1)$th iteration. An illustration of this property is shown in Figure 1. Our experimental results in Section V confirm that, the smaller the AoD $\phi_m^{(i)}$ is, the more likely the algorithm can identify a correct neighbor in the $(m+1)$th iteration, overall leading to higher global subspace data clustering accuracy.

3. To further characterize the AoD of MP and OMP iterations, let $\mathcal{N}_{m-1} \triangleq \{\mathbf{y}_{i_1}, \mathbf{y}_{i_2}, \cdots, \mathbf{y}_{i_{m-1}}\}$ be the set of neighbors selected during the first $m-1$ rounds. Recall from Table I that, for the MP algorithm, some neighbor points in $\mathcal{N}_{m-1}$ could be the same, and the residual $\mathbf{r}_m^{(i)}$ in the $m$th iteration is orthogonal to just the latest selected neighbor $\mathbf{y}_{i_{m-1}}$. In contrast, for OMP, all recovered neighbors in $\mathcal{N}_{m-1}$ are distinct, and $\mathbf{r}_m^{(i)}$ is orthogonal to the span of $\mathcal{N}_{m-1}$. We can accordingly observe the following.

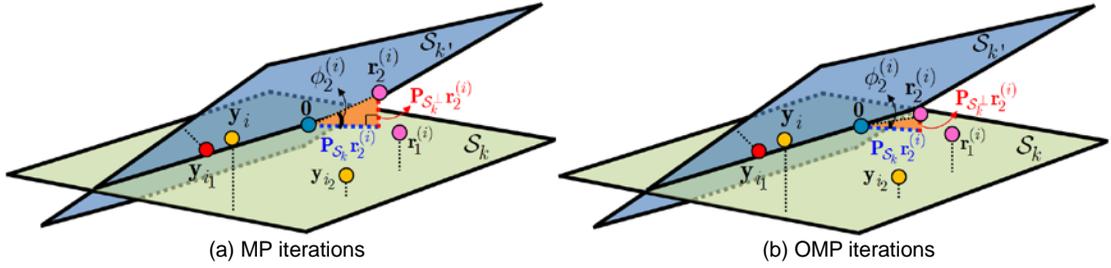

Fig. 3. An illustrative example of Observation 3-2, with ambient space $\mathbb{R}^3$, two subspaces (two-dimensional) $\mathcal{S}_k$ and $\mathcal{S}_{k'}$ close to each other so that SDP does not hold, i.e., $\{\mathbf{y}_{i_1}, \mathbf{y}_{i_2}\} \not\subset \mathcal{S}_k$ (for ease of illustration, only the first two iterations are considered, i.e., $m = 2$). Due to the presence of an incorrect neighbor $\mathbf{y}_{i_1}$ (marked in red), $\mathbf{P}_{\mathcal{S}_k^\perp} \mathbf{r}_2^{(i)}$ for MP (left figure) is larger than that of OMP (right figure), because $\mathbf{P}_{\mathcal{S}_k^\perp} \mathbf{r}_2^{(i)}$ for OMP is required to be orthogonal to both the incorrect neighbor $\mathbf{y}_{i_1}$ and correct neighbor $\mathbf{y}_{i_2}$. Since $\mathbf{P}_{\mathcal{S}_k} \mathbf{r}_2^{(i)}$ of both algorithms are quite close, the AoD $\phi_2^{(i)}$ of the residual vector $\mathbf{r}_2^{(i)}$ for OMP is smaller.

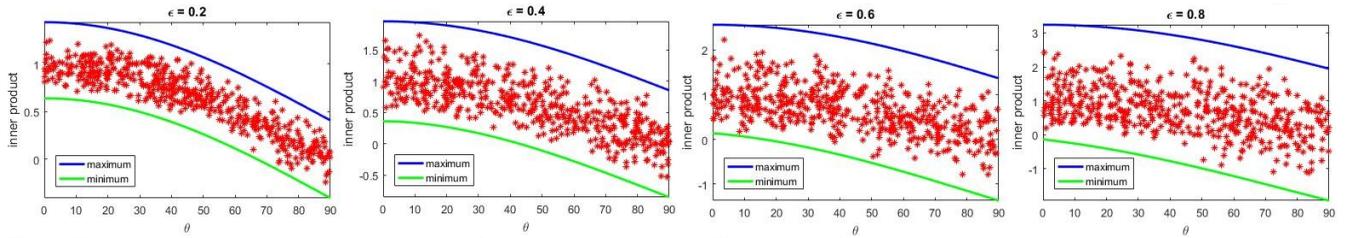

Fig. 4. Illustration of the maximum and minimum of noisy inner product at different noise level $\varepsilon$.

*Observation 3-1:* When the subspaces are well-separated from each other and noise is not large, SDP is likely true so that $\mathcal{N}_{m-1} \subset \mathcal{Y}_k$. In this case, OMP jointly removes from $\mathbf{r}_{m-1}^{(i)}$ the contribution due to $\mathbf{y}_{i_1}, \cdots, \mathbf{y}_{i_{m-1}}$, resulting in a smaller $\mathbf{P}_{\mathcal{S}_k} \mathbf{r}_m^{(i)}$, i.e., the orthogonal projection of $\mathbf{r}_m^{(i)}$ onto the desired subspace $\mathcal{S}_k$, as compared to MP. Meanwhile, since $\mathcal{N}_{m-1} \subset \mathcal{Y}_k$, $\mathbf{P}_{\mathcal{S}_k^\perp} \mathbf{r}_m^{(i)}$ stays largely intact for both MP and OMP. As a result, MP is expected to yield a smaller $\phi_m^{(i)} = \tan^{-1} \left( \left\| \mathbf{P}_{\mathcal{S}_k^\perp} \mathbf{r}_m^{(i)} \right\|_2 / \left\| \mathbf{P}_{\mathcal{S}_k} \mathbf{r}_m^{(i)} \right\|_2 \right)$ (see Figure 2 for an illustration) and therefore a better clustering performance.

*Observation 3-2:* When the subspaces get close to each other, SDP would often be violated so that $\mathcal{N}_{m-1} \not\subset \mathcal{Y}_k$. Still, OMP upon joint orthogonal projection produces smaller $\mathbf{P}_{\mathcal{S}_k} \mathbf{r}_m^{(i)}$ as compared to MP. However, the presence of incorrect neighbors causes different impacts on $\mathbf{P}_{\mathcal{S}_k^\perp} \mathbf{r}_m^{(i)}$. Indeed, since the residual $\mathbf{r}_m^{(i)}$ of OMP is required to be orthogonal to *all* the already selected data points $\mathbf{y}_{i_1}, \cdots, \mathbf{y}_{i_{m-1}}$, in particular, those in-correct neighbors, this will also remove from $\mathbf{r}_m^{(i)}$ a somewhat significant component in $\mathcal{S}_k^\perp$, leading to a far smaller $\mathbf{P}_{\mathcal{S}_k^\perp} \mathbf{r}_m^{(i)}$ as compared to MP. As such, OMP is expected to yield a smaller AoD (see Figure 3 for an illustration) and, in turn, achieve higher data clustering accuracy as compared to MP. Our experimental results in Section V confirm these observations.

4. We shall notice the marked difference between the results in the probabilistic framework [19] and our study. Indeed, the sufficient conditions for SDP in [19] are characterized using *global* geometric features, such as subspace affinity and sample density, and are concerned with the requirements on the ground truth subspaces that ensure correct neighbor recovery. On the contrary, our results in Theorem 3.2 manifest the required orientation of the residual vectors using inter/intra-cluster coherence, in-radius, and minimal angles, mostly related to the *local* geometry of the given data set. In particular, Theorem 3.2 and the above discussions explicitly reveal that the AoD of the residual vector plays a pivotal role in noisy SSC using greedy-based neighbor identification; such a conclusion cannot be deduced from the subspace affinity based sufficient conditions in [19].

5. Finally, we remark on Lemma 3.1. Since the function $f_\varphi$ defined in (6) is continuous on the compact set $[0, 2\pi]$, the extreme values in (7) and (8) exist [20]. It can be shown (see Appendix) that, for a given $\varphi$, the maximum/minimum of $f_\varphi$ can be obtained by solving the roots of a polynomial of order 4. Hence, the ex-treme values $\max_{\theta \in [0,2\pi]} f_{\cos^{-1}(\mu_c(\mathcal{X}_k))}(\theta)$ and $\min_{\theta \in [0,2\pi]} f_{\cos^{-1}(\eta_k)}(\theta)$ involved in the sufficient condition (9) can be obtained using numerically efficient solvers (actually, the closed-form solutions exist, though the exact forms are rather complicated). Below we use numerical simulations to corroborate the analytical results in Lemma 3.1. We consider an ambient signal space dimension $n = 5$, and generate 5000 unit-norm vectors according to

$$\begin{cases} \mathbf{x}_1 = [\cos(\varphi/2) \ \sin(\varphi/2) \ \mathbf{0}^T]^T \\ \mathbf{x}_2 = [\cos(\varphi/2) \ -\sin(\varphi/2) \ \mathbf{0}^T]^T, \end{cases} \quad (12)$$

where $\varphi$ is uniformly generated from the open interval $(0, \pi/2)$. Afterwards, 5000 independent noise vector pairs $(\mathbf{e}_1, \mathbf{e}_2)$ are drawn uniformly from the closed $\varepsilon$-ball, and the respective inner products $(\mathbf{x}_1 + \mathbf{e}_1)^T(\mathbf{x}_2 + \mathbf{e}_2)$ are computed. Figure 4 plots the obtained inner products (red points), together with the theoretical maximum (blue line) and minimum (green line), as a function of $\theta$ for four different noise levels $\varepsilon = 0.2, \ 0.4, \ 0.6, \ 0.8$. The figure shows that our theoretical results well predict the simulated outcomes.

## 4 MATHEMATICAL PROOFS

### 4.1 Proof of Lemma 3.1

An optimal noise pair $(\overline{\mathbf{e}}_i, \overline{\mathbf{e}}_j)$ that achieves either maximum or minimum must be of norm equal to $\varepsilon$. To see this, assume $\|\overline{\mathbf{e}}_i\|_2 = \delta < \varepsilon$. Then, choose $\tilde{\mathbf{e}}_i = \overline{\mathbf{e}}_i \pm (\varepsilon - \delta)\frac{\mathbf{x}_j + \overline{\mathbf{e}}_j}{\|\mathbf{x}_j + \overline{\mathbf{e}}_j\|_2}$, and we have

$$\begin{aligned} \|\tilde{\mathbf{e}}_i\|_2 &= \left\| \overline{\mathbf{e}}_i \pm (\varepsilon - \delta)\frac{\mathbf{x}_j + \overline{\mathbf{e}}_j}{\|\mathbf{x}_j + \overline{\mathbf{e}}_j\|_2} \right\|_2 \\ &\leq \|\overline{\mathbf{e}}_i\|_2 + \left\| (\varepsilon - \delta)\frac{\mathbf{x}_j + \overline{\mathbf{e}}_j}{\|\mathbf{x}_j + \overline{\mathbf{e}}_j\|_2} \right\|_2 = \delta + (\varepsilon - \delta) = \varepsilon, \end{aligned} \quad (13)$$

such that

$$\begin{aligned} (\mathbf{x}_i + \tilde{\mathbf{e}}_i)^T(\mathbf{x}_j + \overline{\mathbf{e}}_j) &= (\mathbf{x}_i + \overline{\mathbf{e}}_i \pm (\varepsilon - \delta)\frac{\mathbf{x}_j + \overline{\mathbf{e}}_j}{\|\mathbf{x}_j + \overline{\mathbf{e}}_j\|_2})^T(\mathbf{x}_j + \overline{\mathbf{e}}_j) \\ &= (\mathbf{x}_i + \overline{\mathbf{e}}_i)^T(\mathbf{x}_j + \overline{\mathbf{e}}_j) \pm (\varepsilon - \delta)\|\mathbf{x}_j + \overline{\mathbf{e}}_j\|_2, \end{aligned} \quad (14)$$

implying that $(\mathbf{x}_i + \overline{\mathbf{e}}_i)^T(\mathbf{x}_j + \overline{\mathbf{e}}_j)$ attains neither maximum nor minimum. We go on to show that such a noise pair $(\overline{\mathbf{e}}_i, \overline{\mathbf{e}}_j)$ must belong to $Span\{\mathbf{x}_i, \mathbf{x}_j\}$. To verify this, consider the associated Lagrangian

$$\begin{aligned} \mathcal{L}(\mathbf{e}_i, \mathbf{e}_j, \lambda_1, \lambda_2) \\ = (\mathbf{x}_i + \mathbf{e}_i)^T(\mathbf{x}_j + \mathbf{e}_j) + \lambda_1(\mathbf{e}_i^T\mathbf{e}_i - \varepsilon^2) + \lambda_2(\mathbf{e}_j^T\mathbf{e}_j - \varepsilon^2). \end{aligned} \quad (15)$$

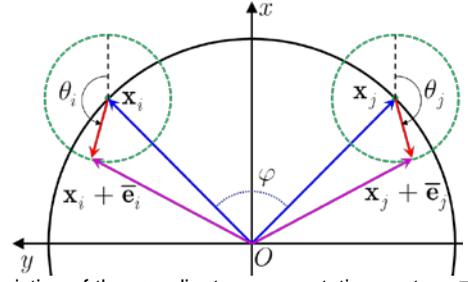

Fig. 5. Depiction of the coordinate representation vectors $\overline{\mathbf{e}}_i$, $\overline{\mathbf{e}}_j$, $\mathbf{x}_i$, and $\mathbf{x}_j$ on the xy-plane, on which $\mathbf{x}_i$ and $\mathbf{x}_j$ are symmetric with respect to the x-axis.

Based on (15), the first-order necessary condition $\nabla_{\mathbf{e}_i}\mathcal{L}(\mathbf{e}_i, \mathbf{e}_j, \lambda_1, \lambda_2) = 0$ and $\nabla_{\mathbf{e}_j}\mathcal{L}(\mathbf{e}_i, \mathbf{e}_j, \lambda_1, \lambda_2) = 0$ are given by, respectively, $\mathbf{x}_j + \mathbf{e}_j + 2\lambda_1\mathbf{e}_i = 0$ and $\mathbf{x}_i + \mathbf{e}_i + 2\lambda_2\mathbf{e}_j = 0$, which in a matrix form reads

$$\begin{bmatrix} \mathbf{e}_i & \mathbf{e}_j \end{bmatrix} \underbrace{\begin{bmatrix} 2\lambda_1 & 1 \\ 1 & 2\lambda_2 \end{bmatrix}}_{\mathbf{A}} = -\begin{bmatrix} \mathbf{x}_j & \mathbf{x}_i \end{bmatrix}. \quad (16)$$

If $\mathbf{A}$ is full rank, (16) implies

$$\begin{bmatrix} \mathbf{e}_i & \mathbf{e}_j \end{bmatrix} = -\begin{bmatrix} \mathbf{x}_j & \mathbf{x}_i \end{bmatrix}\mathbf{A}^{-1}, \quad (17)$$

confirming that $\overline{\mathbf{e}}_i, \overline{\mathbf{e}}_j \in Span\{\mathbf{x}_i, \mathbf{x}_j\}$. The case in which $\mathbf{A}$ is not full rank, thus $4\lambda_1\lambda_2 = 1$, is impossible, and should be precluded. Indeed, if $4\lambda_1\lambda_2 = 1$, (16) becomes

$$\begin{cases} \mathbf{x}_j + 2\lambda_1(\overline{\mathbf{e}}_i + \overline{\mathbf{e}}_j/(2\lambda_1)) = 0 \\ \mathbf{x}_i + (\overline{\mathbf{e}}_i + \overline{\mathbf{e}}_j/(2\lambda_1)) = 0 \end{cases}, \quad (18)$$

which implies that $\mathbf{x}_i$ is parallel to $\mathbf{x}_j$, contracting with the assumption $0 < \cos^{-1}(\mathbf{x}_i^T\mathbf{x}_j) < \pi/2$.

Now since $\overline{\mathbf{e}}_i, \overline{\mathbf{e}}_j \in Span\{\mathbf{x}_i, \mathbf{x}_j\}$, we can focus on the two-dimensional plane $Span\{\mathbf{x}_i, \mathbf{x}_j\}$ in the subsequent analysis. To further ease discussions, let us choose $\{\mathbf{b}_1, \mathbf{b}_2\}$ to be an orthonormal basis for $Span\{\mathbf{x}_i, \mathbf{x}_j\}$, compute accordingly the coordinate representation vectors [21] of $\overline{\mathbf{e}}_i$, $\overline{\mathbf{e}}_j$, $\mathbf{x}_i$, and $\mathbf{x}_j$ with respect to $\{\mathbf{b}_1, \mathbf{b}_2\}$, and then depict all the four coordinate vectors on the two-dimensional xy-plane. Since $\{\mathbf{b}_1, \mathbf{b}_2\}$ is orthonormal, the inner product of any two vectors in $Span\{\mathbf{x}_i, \mathbf{x}_j\} \subset \mathbb{R}^n$ is identical to the inner product of the corresponding coordinate representation vectors on the xy-plane [21]. Without loss of generality, assume $\{\mathbf{b}_1, \mathbf{b}_2\}$ is chosen so that the coordinate representations of $\mathbf{x}_i$ and $\mathbf{x}_2$ are symmetric with respect to the x-axis (as shown in Figure 5). Meanwhile, to conserve notation and without causing confusion, the same symbols $\overline{\mathbf{e}}_i$, $\overline{\mathbf{e}}_j$, $\mathbf{x}_i$, and $\mathbf{x}_j$ are used to denote the corresponding coordinate representation vectors. The proof of Lemma 3.1 relies on the next lemma, whose proof is relegated to the end of this section.

**Lemma 4.1.** *Let $\theta_i$ be the angle, measured counter clock-wise with respect to the x-axis, assumed by the optimal perturbation $\overline{\mathbf{e}}_i$, and $\theta_j$ be the angle, measured clock-wise with respect to the x-axis, for $\overline{\mathbf{e}}_j$. Then we have $\theta_i = \theta_j$.* □

Based on Lemma 4.1, we can assume $\theta_i = \theta_j = \theta$, where $\theta \in [0, 2\pi]$. With some manipulations we have

$$\begin{aligned}(\mathbf{x}_i + \overline{\mathbf{e}}_i)^T(\mathbf{x}_j + \overline{\mathbf{e}}_j) \\ = \mathbf{x}_i^T \mathbf{x}_j + 2\varepsilon \cos\left(\cos^{-1}(\mathbf{x}_i^T \mathbf{x}_j / 2) + \theta\right) + \varepsilon^2 \cos(2\theta) \\ = \mathbf{x}_i^T \mathbf{x}_j + f_{\cos^{-1}(\mathbf{x}_i^T \mathbf{x}_j)}(\theta)\end{aligned} \quad (19)$$

Note that, since $f_{\cos^{-1}(\mathbf{x}_i^T \mathbf{x}_j)}(\theta)$ is continuous on the compact set $[0, 2\pi]$, the maximum and minimum exist. The proof of Lemma 3.1 is completed.

**Proof of Lemma 4.1.** We denote $\varphi = \cos^{-1}(\mathbf{x}_i^T \mathbf{x}_j)$ and write $\overline{\mathbf{e}}_i = \varepsilon[\cos\theta_i \; \sin\theta_i]^T$ and $\overline{\mathbf{e}}_j = \varepsilon[\cos\theta_j \; -\sin\theta_j]^T$. Then with some manipulations we have

$$\begin{aligned}(\mathbf{x}_i + \overline{\mathbf{e}}_i)^T(\mathbf{x}_j + \overline{\mathbf{e}}_j) \\ = \cos\varphi + \varepsilon(\cos(\varphi/2 + \theta_i) + \cos(\varphi/2 + \theta_j)) + \varepsilon^2 \cos(\theta_i + \theta_j).\end{aligned} \quad (20)$$

Based on (20), the 1st-order necessary condition $\frac{\partial (\mathbf{x}_i + \overline{\mathbf{e}}_i)^T(\mathbf{x}_j + \overline{\mathbf{e}}_j)}{\partial \theta_i} = \frac{\partial (\mathbf{x}_i + \overline{\mathbf{e}}_i)^T(\mathbf{x}_j + \overline{\mathbf{e}}_j)}{\partial \theta_j} = 0$ can be expressed as

$$\begin{cases} \sin\left((\varphi/2) + \theta_i\right) + \varepsilon \sin(\theta_i + \theta_j) = 0, \\ \sin\left((\varphi/2) + \theta_j\right) + \varepsilon \sin(\theta_i + \theta_j) = 0, \end{cases} \quad (21)$$

which implies $\sin\left((\varphi/2) + \theta_i\right) = \sin\left((\varphi/2) + \theta_j\right)$, meaning that

$$\theta_i = \theta_j \quad \text{or} \quad \theta_j = (2k+1)\pi - \theta_i \quad k \in \mathbb{Z}. \quad (22)$$

The extreme values of $(\mathbf{x}_i + \overline{\mathbf{e}}_i)^T(\mathbf{x}_j + \overline{\mathbf{e}}_j)$ will not occur when $\theta_j = (2k+1)\pi - \theta_i$, $k \in \mathbb{Z}$, which is thus precluded from in (22). Indeed, if $\theta_j = (2k+1)\pi - \theta_i$, (20) becomes

$$\begin{aligned}(\mathbf{x}_i + \overline{\mathbf{e}}_i)^T(\mathbf{x}_j + \overline{\mathbf{e}}_j) \\ = \cos\varphi + \varepsilon\left(\cos((\varphi/2) + \theta_i) - \cos((\varphi/2) - \theta_i)\right) - \varepsilon^2 \quad (23) \\ = \cos\varphi - 2\varepsilon \sin(\varphi/2)\sin\theta_i - \varepsilon^2.\end{aligned}$$

For $\mathbf{e}'_i = \mathbf{e}'_j = \varepsilon[1 \; 0]^T$ and $\mathbf{e}''_i = \varepsilon[0 \; 1]^T = -\mathbf{e}''_j$, after some manipulations it can be shown that

$$\begin{aligned}(\mathbf{x}_i + \mathbf{e}'_i)^T(\mathbf{x}_j + \mathbf{e}'_j) = \cos\varphi + 2\varepsilon\cos(\varphi/2) + \varepsilon^2 \\ \geq \cos\varphi - 2\varepsilon \sin(\varphi/2)\sin(\theta_i) - \varepsilon^2 = (\mathbf{x}_i + \overline{\mathbf{e}}_i)^T(\mathbf{x}_j + \overline{\mathbf{e}}_j) \\ \geq \cos\varphi - 2\varepsilon \sin(\varphi/2) - \varepsilon^2 = (\mathbf{x}_i + \mathbf{e}''_i)^T(\mathbf{x}_j + \mathbf{e}''_j),\end{aligned} \quad (24)$$

which implies $(\mathbf{x}_i + \overline{\mathbf{e}}_i)^T(\mathbf{x}_j + \overline{\mathbf{e}}_j)$ does not attain the extreme values. The proof is completed. □

### 4.2 Proof of Theorem 3.2

The proof is done by induction. Let us first consider the first iteration ($m = 1$). From Table I, it is easy to see that both MP and OMP can select a data point from the correct cluster $\mathcal{Y}_k$ if the following condition is satisfied:

$$\max_{\mathbf{y}_l \in \mathcal{Y} \setminus \mathcal{Y}_k} \left|\mathbf{y}_i^T \mathbf{y}_l\right| < \max_{\mathbf{y}_j \in \mathcal{Y}_k \setminus \{\mathbf{y}_i\}} \left|\mathbf{y}_i^T \mathbf{y}_j\right|. \quad (25)$$

To obtain a sufficient condition ensuring (25), we shall first derive an upper bound and a lower bound for, respectively, the LHS and RHS of (25). To proceed, we first note in the noiseless case that

$$\begin{cases} \max_{\mathbf{x}_l \in \mathcal{X} \setminus \mathcal{X}_k} |\mathbf{x}_i^T \mathbf{x}_l| \leq \mu_c(\mathcal{X}_k), \\ \max_{\mathbf{x}_j \in \mathcal{X}_k \setminus \{\mathbf{x}_i\}} |\mathbf{x}_i^T \mathbf{x}_j| \overset{(a)}{\geq} r(\mathcal{P}_{-i}^k), \end{cases} \quad (26)$$

where (a) is proved in [17]. The first equation of (26) implies

$$\max_{\mathbf{x}_l \in \mathcal{X} \setminus \mathcal{X}_k, \mathbf{x}_i^T \mathbf{x}_l > 0} \mathbf{x}_i^T \mathbf{x}_l \leq \mu_c(\mathcal{X}_k), \quad (27)$$

and

$$\max_{\mathbf{x}_l \in \mathcal{X} \setminus \mathcal{X}_k, \mathbf{x}_i^T \mathbf{x}_l < 0} \mathbf{x}_i^T(-\mathbf{x}_l) \leq \mu_c(\mathcal{X}_k). \quad (28)$$

From Lemma 3.1 and (27), we have, for $\mathbf{y}_l \in \mathcal{Y} \setminus \mathcal{Y}_k$ such that $\mathbf{x}_i^T \mathbf{x}_l > 0$,

$$\mathbf{y}_i^T \mathbf{y}_l \leq \mu_c(\mathcal{X}_k) + \varepsilon \max_{\theta \in [0, 2\pi]} f_{\cos^{-1}(\mu_c(\mathcal{X}_k))}(\theta); \quad (29)$$

also, Lemma 3.1 together with (28) implies, for $\mathbf{y}_l \in \mathcal{Y} \setminus \mathcal{Y}_k$ such that $\mathbf{x}_i^T \mathbf{x}_l < 0$, we have

$$-\mathbf{y}_i^T \mathbf{y}_l \leq \mu_c(\mathcal{X}_k) + \varepsilon \max_{\theta \in [0, 2\pi]} f_{\cos^{-1}(\mu_c(\mathcal{X}_k))}(\theta). \quad (30)$$

Combining (29) and (30) yields

$$\max_{\mathbf{y}_l \in \mathcal{Y} \setminus \mathcal{Y}_k} \left|\mathbf{y}_i^T \mathbf{y}_l\right| \leq \mu_c(\mathcal{X}_k) + \varepsilon \max_{\theta \in [0, 2\pi]} f_{\cos^{-1}(\mu_c(\mathcal{X}_k))}(\theta). \quad (31)$$

Similarly, based on the second equation of (26) and Lemma 3.1, it can be readily shown that

$$\max_{\mathbf{y}_j \in \mathcal{Y}_k \setminus \{\mathbf{y}_i\}} \left|\mathbf{y}_i^T \mathbf{y}_j\right| \geq r(\mathcal{P}_{-i}^k) + \varepsilon \min_{\theta \in [0, 2\pi]} f_{\cos^{-1}(\eta_k)}(\theta). \quad (32)$$

From (31) and (32), a sufficient condition guaranteeing (25) is thus

$$\begin{aligned}\mu_c(\mathcal{X}_k) + \varepsilon \max_{\theta \in [0, 2\pi]} f_{\cos^{-1}(\mu_c(\mathcal{X}_k))}(\theta) \\ \leq r(\mathcal{P}_{-i}^k) + \varepsilon \min_{\theta \in [0, 2\pi]} f_{\cos^{-1}(\eta_k)}(\theta),\end{aligned} \quad (33)$$

which is true if (9) holds.

Assume that the data points selected during the first $m$ iterations, say, $\mathbf{y}_{i_1}, \mathbf{y}_{i_2}, \cdots, \mathbf{y}_{i_m}$, all belong to the correct cluster $\mathcal{Y}_k$. Then, both MP and OMP can select a point from $\mathcal{Y}_k$ in the $(m+1)$ th iteration provided that

$$\max_{\mathbf{y}_l \in \mathcal{Y} \setminus \mathcal{Y}_k} \left| \mathbf{r}_m^{(i)T} \mathbf{y}_l \right| < \max_{\mathbf{y}_j \in \mathcal{Y}_k \setminus \{\mathbf{y}_i\}} \left| \mathbf{r}_m^{(i)T} \mathbf{y}_j \right|. \tag{34}$$

Still, our goal is to derive a sufficient condition guaranteeing (34). Towards this end, an upper bound for the LHS of (34) can be obtained as follows:

$$\begin{aligned}
\max_{\mathbf{y}_l \in \mathcal{Y} \setminus \mathcal{Y}_k} \left| \mathbf{r}_m^{(i)T} \mathbf{y}_l \right| &= \max_{\mathbf{x}_l \in \mathcal{X} \setminus \mathcal{X}_k} \left| \mathbf{r}_m^{(i)T} (\mathbf{x}_l + \mathbf{e}_l) \right| \\
&\stackrel{(b)}{\leq} \max_{\mathbf{x}_l \in \mathcal{X} \setminus \mathcal{X}_k} \left( \left| \mathbf{r}_m^{(i)T} \mathbf{x}_l \right| + \left| \mathbf{r}_m^{(i)T} \mathbf{e}_l \right| \right) \\
&= \left\| \mathbf{r}_m^{(i)} \right\|_2 \max_{\mathbf{x}_l \in \mathcal{X} \setminus \mathcal{X}_k} \left( \underbrace{\|\mathbf{x}_l\|_2}_{=1} \left| \cos \angle(\mathbf{r}_m^{(i)}, \mathbf{x}_l) \right| + \underbrace{\|\mathbf{e}_l\|_2}_{\leq \varepsilon} \underbrace{\left| \cos \angle(\mathbf{r}_m^{(i)}, \mathbf{e}_l) \right|}_{\leq 1} \right) \\
&\leq \left\| \mathbf{r}_m^{(i)} \right\|_2 \max_{\mathbf{x}_l \in \mathcal{X} \setminus \mathcal{X}_k} \left( \left| \cos \angle(\mathbf{r}_m^{(i)}, \mathbf{x}_l) \right| + \varepsilon \right) \\
&\stackrel{(c)}{\leq} \left\| \mathbf{r}_m^{(i)} \right\|_2 (\cos(\max\{\theta_k - \phi_m^{(i)}, 0\}) + \varepsilon),
\end{aligned} \tag{35}$$

where (b) follows from triangular inequality, and the proof of (c) is given at the end of this section. A lower bound for the RHS of (34) is

$$\begin{aligned}
\max_{\mathbf{y}_j \in \mathcal{Y}_k \setminus \{\mathbf{y}_i\}} \left| \mathbf{r}_m^{(i)T} \mathbf{y}_j \right| &= \max_{\mathbf{y}_j \in \mathcal{Y}_k \setminus \{\mathbf{y}_i\}} \left| \mathbf{r}_m^{(i)T} (\mathbf{x}_j + \mathbf{e}_j) \right| \\
&\stackrel{(d)}{\geq} \max_{\mathbf{y}_j \in \mathcal{Y}_k \setminus \{\mathbf{y}_i\}} \left( \left| \mathbf{r}_m^T \mathbf{x}_j \right| - \left| \mathbf{r}_m^T \mathbf{e}_j \right| \right) \\
&= \left\| \mathbf{r}_m^{(i)} \right\|_2 \max_{\mathbf{y}_j \in \mathcal{Y}_k \setminus \{\mathbf{y}_i\}} \left( \left| \cos \angle(\mathbf{r}_m^{(i)}, \mathbf{x}_j) \right| - \varepsilon \left| \cos \angle(\mathbf{r}_m^{(i)}, \mathbf{e}_j) \right| \right) \\
&\geq \left\| \mathbf{r}_m^{(i)} \right\|_2 \max_{\mathbf{y}_j \in \mathcal{Y}_k \setminus \{\mathbf{y}_i\}} \left( \left| \cos \angle(\mathbf{r}_m^{(i)}, \mathbf{x}_j) \right| - \varepsilon \right),
\end{aligned} \tag{36}$$

where (d) follows from the triangular inequality. Combining (35) and (36), a sufficient condition guaranteeing (34) is immediately obtained as in (10). The proof of theorem is thus completed.

**Proof of (c) in (35).** It suffices to show

$$\max_{\mathbf{x}_l \in \mathcal{X} \setminus \mathcal{X}_k} \left| \cos \angle(\mathbf{r}_m^{(i)}, \mathbf{x}_l) \right| \leq \cos(\theta_k - \phi_m^{(i)}) \text{ if } \theta_k \geq \phi_m^{(i)}. \tag{37}$$

We will prove (37) by showing that

$$\angle(\mathbf{r}_m^{(i)}, \mathbf{x}_l) \geq \theta_k - \phi_m^{(i)} \text{ if } \theta_k \geq \phi_m^{(i)}. \tag{38}$$

By definition of $\theta_k$, for every $\mathbf{x}_j \in \mathcal{S}_k$ and $\mathbf{x}_l \in \mathcal{S}_{\bar{k}}$ with $\bar{k} \neq k$, we have

$$\angle(\mathbf{x}_j, \mathbf{x}_l) \geq \theta_k. \tag{39}$$

Note that, since both $\mathbf{x}_j$ and $\mathbf{x}_l$ are unit-norm, the angle $\angle(\mathbf{x}_j, \mathbf{x}_l)$ is equal to the shortest arc length $\widehat{\mathbf{x}_j \mathbf{x}_l}$ on the unit-sphere that connects $\mathbf{x}_j$ and $\mathbf{x}_l$, and (39) is thus equivalent to

$$\widehat{\mathbf{x}_j \mathbf{x}_l} \geq \theta_k. \tag{40}$$

Since $\left\| \mathbf{P}_{\mathcal{S}_k^\perp} \mathbf{r}_m^{(i)} \right\|_2 / \left\| \mathbf{P}_{\mathcal{S}_k} \mathbf{r}_m^{(i)} \right\|_2 = \tan(\phi_m^{(i)})$, the unit-norm vector

$$\mathbf{w} = \mathbf{P}_{\mathcal{S}_k} \mathbf{r}_m^{(i)} / \left\| \mathbf{P}_{\mathcal{S}_k} \mathbf{r}_m^{(i)} \right\|_2 \in \mathcal{S}_k \tag{41}$$

yields

$$\begin{aligned}
\angle(\mathbf{w}, \mathbf{r}_m^{(i)}) &= \cos^{-1}(\mathbf{w}^T \mathbf{r}_m^{(i)} / \left\| \mathbf{r}_m^{(i)} \right\|_2) \\
&= \cos^{-1}(\left\| \mathbf{P}_{\mathcal{S}_k} \mathbf{r}_m^{(i)} \right\|_2 / \left\| \mathbf{r}_m^{(i)} \right\|_2) = \phi_m^{(i)}.
\end{aligned} \tag{42}$$

For the unit-norm vector $\bar{\mathbf{r}}_m^{(i)} = \mathbf{r}_m^{(i)} / \left\| \mathbf{r}_m^{(i)} \right\|_2$, similar to (40) we have

$$\widehat{\mathbf{w} \bar{\mathbf{r}}_m^{(i)}} = \angle(\mathbf{w}, \bar{\mathbf{r}}_m^{(i)}) = \angle(\mathbf{w}, \mathbf{r}_m^{(i)}) = \phi_m^{(i)}. \tag{43}$$

Now, with (40) and (43), it follows that

$$\widehat{\bar{\mathbf{r}}_m^{(i)} \mathbf{x}_l} + \phi_m^{(i)} = \widehat{\bar{\mathbf{r}}_m^{(i)} \mathbf{x}_l} + \widehat{\mathbf{w} \bar{\mathbf{r}}_m^{(i)}} \stackrel{(e)}{\geq} \widehat{\mathbf{w} \mathbf{x}_l} \geq \theta_k, \tag{44}$$

where (e) is obtained by using spherical triangle inequality [22]. From (44), we then have

$$\angle(\mathbf{r}_m^{(i)}, \mathbf{x}_l) = \angle(\bar{\mathbf{r}}_m^{(i)}, \mathbf{x}_l) = \widehat{\bar{\mathbf{r}}_m^{(i)} \mathbf{x}_l} \geq \theta_k - \phi_m^{(i)} \text{ for } \theta_k \geq \phi_m^{(i)}, \tag{45}$$

which completes the proof. □

## 5 EXPERIMENTAL RESULTS

In this section, experimental results based on both synthetic data and real human face data are provided to validate our analytic study.

### 5.1 Synthetic Data

For the experiments with synthetic data, we consider a ground truth of $L=3$ subspaces in $\mathbb{R}^{100}$, all of an identical dimension $d=20$; the number of samples per subspace is the same, equal to $N_l = 150$, $1 \leq l \leq 3$. The subspace affinity[1], denoted by $\rho_{i,j}$, is used as the metric for gauging the separation between subspaces $\mathcal{S}_i$ and $\mathcal{S}_j$; for ease of illustration, we simply choose $\rho_{i,j} = \rho$, that is, all subspaces are equally separated from each other. The noise-free unit-norm data vectors in $\mathcal{X}_l$ are generated uniformly from the unit-sphere of the $l$th subspace, $1 \leq l \leq 3$; the noisy data cluster $\mathcal{Y}_l$ is obtained by adding to each vector $\mathbf{x}_i$ in $\mathcal{X}_l$ a noise perturbation vector $\mathbf{e}_i$ uniformly drawn from the closed $\varepsilon$-ball in $\mathbb{R}^{100}$. We use both MP and OMP for neighbor identification. For the data point $\mathbf{y}_i$, the residual vector $\mathbf{r}_m^{(i)}$ and the AoD $\phi_m^{(i)}$ in the $m$th iteration are obtained. Associated with each $m$, we compute the average of $\left\| \mathbf{P}_{\mathcal{S}_k} \mathbf{r}_m^{(i)} \right\|_2$ and $\left\| \mathbf{P}_{\mathcal{S}_k^\perp} \mathbf{r}_m^{(i)} \right\|_2$ over all data points, defined to be

$$\bar{r}_m^\| \triangleq \frac{1}{LN_l} \sum_{i=1}^{LN_l} \left\| \mathbf{P}_{\mathcal{S}_k} \mathbf{r}_m^{(i)} \right\|_2 \text{ and } \bar{r}_m^\perp \triangleq \frac{1}{LN_l} \sum_{i=1}^{LN_l} \left\| \mathbf{P}_{\mathcal{S}_k^\perp} \mathbf{r}_m^{(i)} \right\|_2 \tag{46}$$

---

1. The affinity between subspaces $\mathcal{S}_i$ and $\mathcal{S}_j$ is defined to be $\rho(i,j) \triangleq \|\mathbf{U}_i^T \mathbf{U}_j\|_F / \sqrt{\min(d_i, d_j)}$ [19], where the columns of $\mathbf{U}_i$ ($\mathbf{U}_j$, respectively) consist of an orthonormal basis for $\mathcal{S}_i$ ($\mathcal{S}_j$, respectively), and $\|\cdot\|_F$ denotes the matrix Frobenius norm.

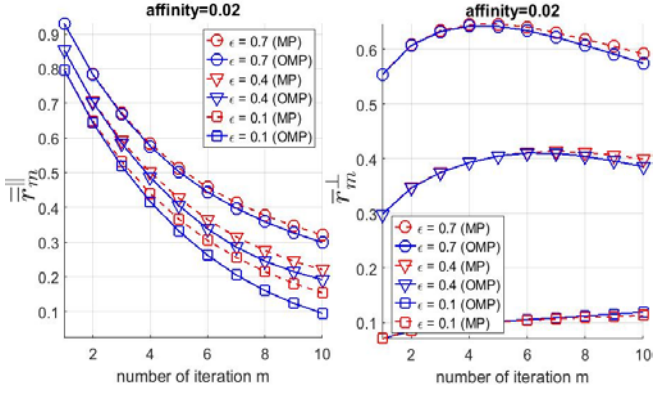

Fig. 6. Illustration of $\bar{r}_m^{\parallel}$ and $\bar{r}_m^{\perp}$ versus the number $m$ of iteration with affinity=0.02.

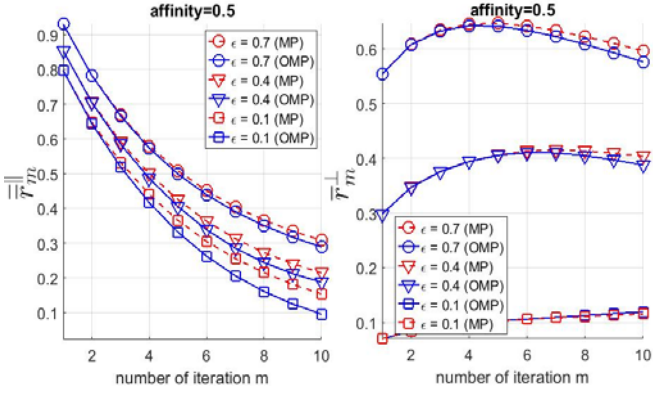

Fig. 7. Illustration of $\bar{r}_m^{\parallel}$ and $\bar{r}_m^{\perp}$ versus the number $m$ of iteration with affinity=0.5.

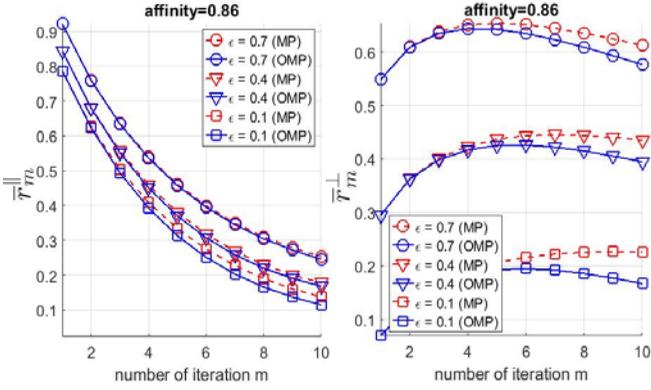

Fig. 8. Illustration of $\bar{r}_m^{\parallel}$ and $\bar{r}_m^{\perp}$ versus the number $m$ of iteration with affinity=0.86.

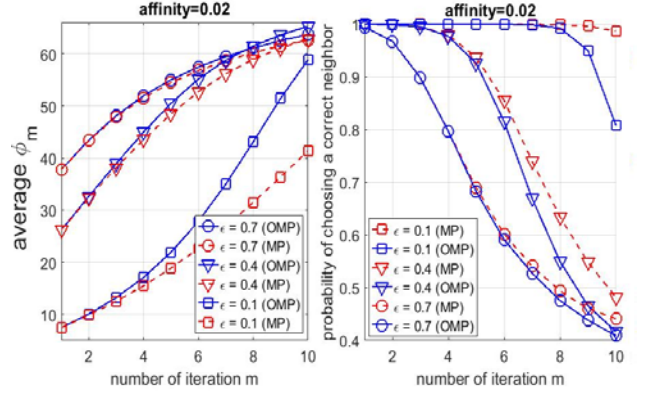

Fig. 9. Average deviation angle $\phi_m$ and probability $P_m$ versus the number $m$ of iteration with affinity=0.02.

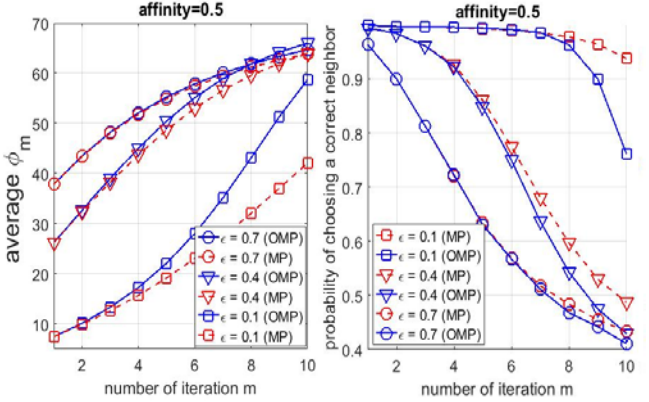

Fig. 10. Average deviation angle $\phi_m$ and probability $P_m$ versus the number $m$ of iteration with affinity=0.5.

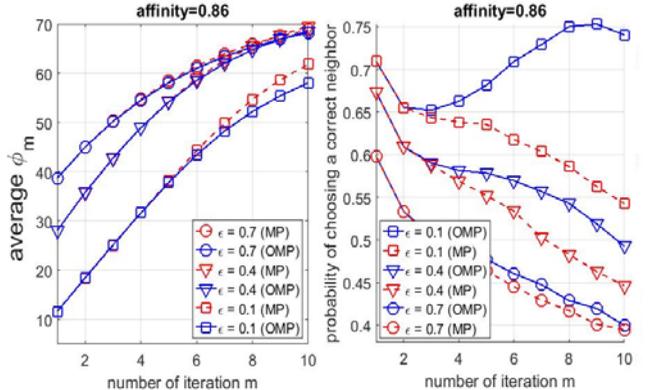

Fig. 11. Average deviation angle $\phi_m$ and probability $P_m$ versus the number $m$ of iteration with affinity=0.86.

and the average of $\phi_m^{(i)}$ as

$$\phi_m \triangleq \frac{1}{LN_l} \sum_{i=1}^{LN_l} \phi_m^{(i)}. \tag{47}$$

As the global data clustering performance measure, we consider the correct clustering rate (CCR), defined to be

$$CCR \triangleq (\# \text{ of correctly classified data points}) / LN_l. \tag{48}$$

We first illustrate the impact of noise on the residual vector $\mathbf{r}_m^{(i)}$ in each iteration. For different subspace affinity, Figures 6~8 show $\bar{r}_m^{\parallel}$ and $\bar{r}_m^{\perp}$ versus $m$; in each figure, three curves corresponding to different noise levels $\varepsilon = 0.1,\ 0.4$, and $0.7$ are plotted. For small to medium subspace affinity (i.e., subspaces fairly well separated away from each other), it is seen from Figures 6 and 7 that, for each $\varepsilon$, (i) OMP entails a noticeable smaller $\bar{r}_m^{\parallel}$, especially when noise is small, (ii) the values of $\bar{r}_m^{\perp}$ achieved by both algorithms are quite close. For large subspace affinity (subspaces close to each other), it can be seen from Figure 8 that (i) still, OMP results in a smaller $\bar{r}_m^{\parallel}$, but the gap between OMP and MP is reduced as compared to that in Figures 6 and 7, (ii) OMP entails a noticeable smaller $\bar{r}_m^{\perp}$ irrespective of the noise levels. All the above findings agree with our discussions and observations made in Section III (see discussion 3). Associated with each subspace affinity, Figures 9~11 accordingly plot $\phi_m$ versus $m$, and the probability that a correct neighbor is selected in the $m$th iteration (denoted by $P_m$). The results show that the averaged $\phi_m$ increases with $\varepsilon$; this is expected since large noise causes large deviation of the residual $\mathbf{r}_m^{(i)}$ from the true subspace $\mathcal{S}_k$. For small to medium subspace affinity, it is seen from Figures 9 and 10 that MP-based search yields smaller averaged $\phi_m$ irrespective of the noise level $\varepsilon$, and is better able to identify a correct neighbor during each iteration (larger $P_m$). This reflects the experimental results of the residual vectors illustrated in Figures 6 and 7, and confirms our discussions in Section III. On the contrary, for large subspace affinity, it is observed from Figure 11 that OMP iterations yield smaller averaged $\phi_m$ and achieve better neighbor identification accuracy (large $P_m$). This is a consequence of our findings in Figure 8, and again supports our discussions in Section III. Figure 12 then plots the CCR curves at different signal-to-noise ratio (SNR), defined to be $10\log(1/\varepsilon^2)$ (in dB); for better illustration, a zoom-in plot with the two curves corresponding to $\rho = 0.86$ removed is shown in Figure 13. The results indicate that, for small to medium subspace affinity, MP achieves higher global data clustering accuracy as compared to OMP. However, for large subspace affinity, OMP outperforms MP because it entails a smaller average AoD as seen in Figure 11.

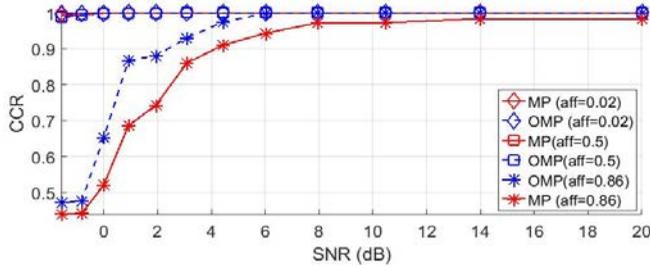

Fig. 12. CCR versus SNR curves of MP and OMP for different subspace affinity.

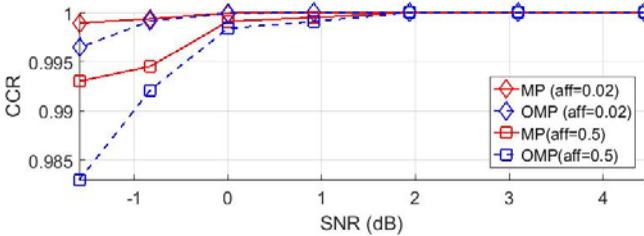

Fig. 13. CCR versus SNR curves of MP and OMP for different subspace affinity (zoom-in plot of Fig. 12 for small-to-medium affinity).

### 5.2 Extended Yale B Data

We go on to test the performances of MP- and OMP-based SSC using the Extended Yale B human face data set as in [8], which consists of 38 human faces (subspaces), each containing 64 data points. To make the computation feasible, before conducting data segmentation we adopt the same technique as in [8] to reduce the ambient space dimension from 2016 to 200. In each trial, we randomly choose $L$ human faces (subspaces) from a total of 38 clusters. For different numbers of subspaces $L = 3, 4, 5$, Figure 14 plots $\overline{r}_m^{\parallel}$ and $\overline{r}_m^{\perp}$ versus $m$, and Figure 15 plots $\phi_m$ and $P_m$ versus $m$. As the figures show, MP yields a smaller $\phi_m$ and a larger $P_m$. Figure 16 then depicts the CCR curves, showing that MP achieves larger CCR. All these results are consistent with the previous findings deduced from the synthetic-data experiments, and again support our analyses and discussions in Section 3.

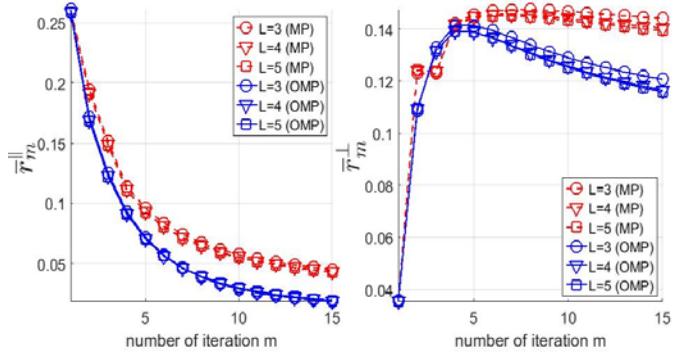

Fig. 14. Illustration of $\overline{r}_m^{\parallel}$ and $\overline{r}_m^{\perp}$ versus the number $m$ of iteration (Extended Yale B human face data).

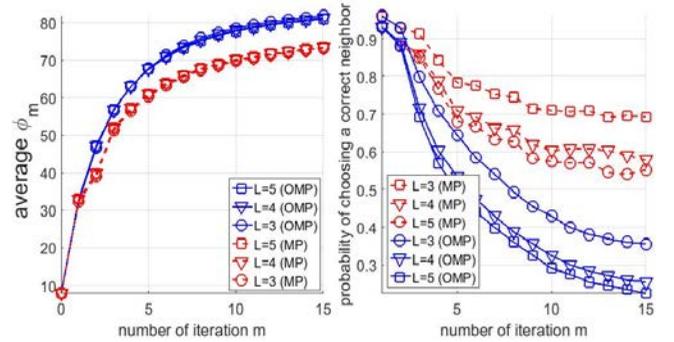

Fig. 15. Average deviation angle $\phi_m$ and probability $P_m$ versus the number $m$ of iteration (Extended Yale B human face data).

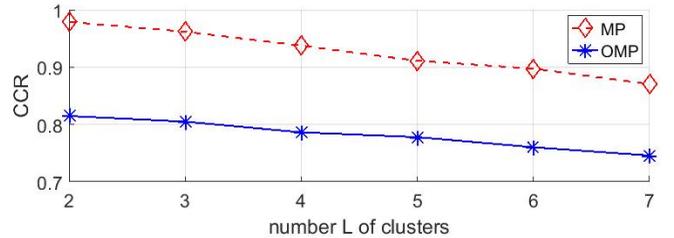

Fig. 16. CCR versus number $L$ of subspaces (people) (Extended Yale B human face data).

## 6 CONCLUSION

Under the deterministic bounded noise corruption, we derive coherence-based sufficient conditions guaranteeing correct neighbor identification for SSC using

MP/OMP. Our study extends existing noiseless coherence-based conditions in [17-18] to the noisy case; moreover, it offers different insights into neighbor recovery under noise as compared to the probabilistic framework [19] using subspace affinity. Our analytic results show that the noise-incurred deviation of the residual vector from the desired subspace, measured by AoD, plays a key role for the neighbor identification performance. If noise is so small that the incurred AoD is sufficiently small (characterized by an inequality in terms of AoD, the minimal angles between subspaces, inter-cluster coherence, and in-radius), correct neighbor recovery is provably true. In particular, for well-separated subspaces, MP is seen to yield smaller AoD, thereby better able to identify correct neighbors and, in turn, achieving higher global data clustering accuracy. Extensive numerical studies using both synthetic and real human face data are used to validate the obtained analytic study. Since MP is more computationally efficient than OMP, our study concludes that MP-based neighbor identification is preferred for fairly well subspace orientation.

## APPENDIX

By definition $f_\varphi$ in (6), the 1st-order necessary condition of the optimization problems involved in (7) and (8) reads

$$\sin((\varphi/2) + \theta) + \varepsilon \sin(2\theta) = 0. \quad (A.1)$$

Through a change of variables with

$$\begin{cases} A_1 = ((2\theta) + (\varphi/2 + \theta))/2 = (3\theta/2) + (\varphi/4), \\ A_2 = ((2\theta) - (\varphi/2 + \theta))/2 = (\theta/2) - (\varphi/4), \end{cases} \quad (A.2)$$

equation (A.1) can be rewritten as

$$\sin(A_1 - A_2) + \varepsilon \sin(A_1 + A_2) = 0. \quad (A.3)$$

Using the angle sum and difference identities [23] and with some manipulations, (A.3) becomes

$$(1+\varepsilon)\tan A_1 = (1-\varepsilon)\tan A_2. \quad (A.4)$$

Note that (A.2) implies

$$A_1 = 3A_2 + \varphi. \quad (A.5)$$

Using (A.5) and the angle sum identities [23], (A.4) becomes

$$\frac{\tan 3A_2 + \tan \varphi}{1 - \tan 3A_2 \tan \varphi} = \frac{1-\varepsilon}{1+\varepsilon}\tan A_2, \quad (A.6)$$

which based on the triple angle identity [23] can be further rearranged into

$$\frac{3\tan A_2 - \tan^3 A_2 + \tan \varphi \cdot (1 - 3\tan^2 A_2)}{1 - 3\tan^2 A_2 - \tan \varphi \cdot (3\tan A_2 - \tan^3 A_2)} = E \tan A_2, \quad (A.7)$$

where $E \triangleq (1-\varepsilon)/(1+\varepsilon)$. Finally, after some manipulations, (A.7) admits the following expression:

$$g(\tan A_2) = 0, \quad (A.8)$$

where

$$\begin{aligned} g(x) \triangleq\ & E \tan\varphi \cdot x^4 + (1 - 3E) \cdot x^3 \\ & + 3\tan\varphi(1 - E) \cdot x^2 + (E - 3) \cdot x - \tan\varphi \end{aligned} \quad \text{(A.9)}$$

is a polynomial of order 4. Hence, we conclude that the maximizer and minimizer of $f_\varphi$ satisfy

$$g(\tan((\theta/2) - (\varphi/4))) = 0 . \quad \text{(A.10)}$$

One can obtain the maximizer/minimizer by solving the roots of the polynomial $g(x)$ in (A.9) followed by the inverse tangent function operation.